%% file: Formatting-Instructions-LaTeX-2026.tex
\documentclass[letterpaper]{article} 
\usepackage{aaai2026}  
\usepackage{times}  
\usepackage{helvet}  
\usepackage{courier}  
\usepackage[hyphens]{url}  
\usepackage{graphicx} 
\usepackage{natbib}  
\usepackage{caption} 
\usepackage{xcolor}
\usepackage{tcolorbox}
\definecolor{verylightgray}{rgb}{.97,.97,.97}
\usepackage{algorithm}
\usepackage{algorithmic}

\usepackage{amssymb}
\usepackage{amsmath}

\usepackage{multirow}
\usepackage{booktabs}
\usepackage{array}

\newcommand{\PreserveBackslash}[1]{\let\temp=\\#1\let\\=\temp}

\newcolumntype{C}[1]{>{\PreserveBackslash\centering}p{#1}}
\newcolumntype{R}[1]{>{\PreserveBackslash\raggedleft}p{#1}}
\newcolumntype{L}[1]{>{\PreserveBackslash\raggedright}p{#1}}

\usepackage{tikz}
\usepackage{pgfplots}
\pgfplotsset{compat=1.18}

\usepackage{newfloat}
\usepackage{listings}
\DeclareCaptionStyle{ruled}{labelfont=normalfont,labelsep=colon,strut=off} 
\floatstyle{ruled}
\newfloat{listing}{tb}{lst}{}
\floatname{listing}{Listing}
\nocopyright

\title{Attention2Probability: Attention-Driven Terminology Probability Estimation for Robust Speech-to-Text System}
\author{
    Yangfan Du\textsuperscript{\rm 1}\thanks{Work done during internship at ByteDance.}, Jun Zhang\textsuperscript{\rm 2}, Bin Wang\textsuperscript{\rm 2}, Jin Qiu\textsuperscript{\rm 2}, Lu Huang\textsuperscript{\rm 2}, Yuan Ge\textsuperscript{\rm 1}, Xiaoqian Liu\textsuperscript{\rm 1}, Tong Xiao\textsuperscript{\rm 1,3}\equalcontrib, Jingbo Zhu\textsuperscript{\rm 1,3}
}
\affiliations{
    \textsuperscript{\rm 1}School of Computer Science and Engineering, Northeastern University, Shenyang, China\\

    \textsuperscript{\rm 2}ByteDance\\
    \textsuperscript{\rm 3}NiuTrans Research, Shenyang, China\\
    duyangfan\_neu@outlook.com xiaotong@mail.neu.edu.cn
}

\usepackage{bibentry}

\begin{document}

\maketitle

\begin{abstract}
Recent advances in speech large language models (SLMs) have improved speech recognition and translation in general domains, but accurately generating domain-specific terms or neologisms remains challenging. To address this, we propose Attention2Probability: attention-driven terminology probability estimation for robust speech-to-text system, which is lightweight, flexible, and accurate. Attention2Probability converts cross-attention weights between speech and terminology into presence probabilities, and it further employs curriculum learning to enhance retrieval accuracy. 
Furthermore, to tackle the lack of data for speech-to-text tasks with terminology intervention, we create and release a new speech dataset with terminology to support future research in this area.
Experimental results show that Attention2Probability significantly outperforms the VectorDB method on our test set. Specifically, its maximum recall rates reach 92.57\% for Chinese and 86.83\% for English. This high recall is achieved with a latency of only 8.71ms per query.
Intervening in SLMs’ recognition and translation tasks using Attention2Probability-retrieved terms improves terminology accuracy by 6–17\%, while revealing that the current utilization of terminology by SLMs has limitations.
\end{abstract}

\begin{links}
    \link{Code, Data, Models}{https://github.com/bytedance/Attention2Probability}
\end{links}

\section{Introduction}
Automatic Speech Recognition (ASR) \cite{kim2017joint} and Speech Translation (ST) \cite{xu2021stacked} are vital for real-world applications like meeting transcription and simultaneous interpretation. Advances in Speech Large language Models (SLMs) have significantly improved their performance in general scenarios such as daily conversations \cite{tang2023salmonn,chu2023qwen}. However, in specialized domains such as medicine or gaming, SLMs often struggle to generate accurate technical terminology \cite{yang2024promptasr}. 

A straightforward strategy involves leveraging collected terminological datasets to fine-tune SLMs, thereby enabling them to acquire domain-specific vocabulary \cite{chen2024salm}. However, terminology evolves rapidly as new terms emerge and old ones become obsolete daily \cite{coupland2014language, morgan2025hermeneutical}. Fine-tuning is time-consuming, and data collection is difficult, making this approach poorly suited for dynamic terminology intervention.

With SLMs’ in-context learning (ICL) capabilities, new methods avoid fine-tuning by incorporating potential terms into prompts to enhance terminology generation accuracy \cite{lakomkin2024end,gong2024contextual}. ICL demonstrates notable efficacy in terminology intervention but exhibits high prompt sensitivity: accurate candidate terms can serve as references for SLMs, thereby enhancing the accuracy of terminology generation. Consequently, the core challenge lies in retrieving relevant terms.

\input{figs/overview}

To address the retrieval challenge in speech scenarios, recent works such as Seal \cite{sun2025seal} adopt Vector DataBase (VectorDB) for terminology retrieval, inheriting the RAG framework \cite{lewis2020retrieval,gao2023retrieval}. However, unlike text-centric RAG methods, these solutions must incorporate speech-text modal alignment strategies. This adaptation is critical because the inherent modality gap between speech and text precludes direct application of conventional VectorDB techniques \cite{gong2025br}.

However, VectorDB methods face significant issues: 
\begin{itemize}
\item High Training Cost: Training modality alignment models for speech and text requires large-scale datasets and entails extended training durations.
\item Poor Retrieval Accuracy: VectorDB performs retrieval based on cosine similarity, but it's not equivalent to the probability of terms appearing in audio, resulting in suboptimal recall rates.
\end{itemize}

To address these limitations, we propose the Attention2Probability (A2P) method. As shown in Figure \ref{overview}, A2P completely eliminates the VectorDB used in prior approaches, replacing it with a cross-attention mechanism. This mechanism computes attention weights between the speech input and candidate terms, converting these weights directly into the probability of each term's presence in the speech. 
 
Based on these probabilities, the Top-k most probable terms are selected to identify the terms most likely present in the speech. These terms are appended to the prompt, explicitly informing the SLM of likely relevant terms to guide accurate generation of the target-language terminology. Crucially, without requiring any dedicated modal alignment training, A2P achieves high terminology recall accuracy. The main contributions of this paper are summarized as follows:
\begin{itemize}
\item We depart from conventional VectorDB approaches by introducing a cross-attention mechanism for terminology retrieval, providing a new paradigm for future research in this domain.
\item We have constructed a dedicated speech recognition and translation terminological dataset by leveraging MegaTTS \cite{jiang2025megatts} and open-source terminology translation data, and will publicly release it.\
\item  Our proposed A2P framework achieves high terminology recall accuracy.
\item We critically examine limitations in current methods, laying a foundation for further exploration in this area.
\end{itemize}

\section{Related work}
\subsection{Textual Terminology Translation}
For traditional machine translation tasks, achieving terminology translation typically involves two approaches: constrained generation and data augmentation.
Constrained generation methods enforce the model to generate target terminology during the inference stage through a series of constraint mechanisms \cite{hokamp2017lexically,post2018fast,hasler2018neural}. For instance, during beam search, score rewards are assigned to translation candidates containing terminology to boost the ranking of terminology-related candidates.

The data augmentation schemes are further divided into two types: Placeholder and Code-Switch.
The Placeholder method replaces terms in the source text with special tokens, and then replaces these special tokens with target terms after translation \cite{dinu2019training,bergmanis2021facilitating}.
Code-Switch method doesn't use special tokens; instead, it directly replaces source-language terms with target terms \cite{crego2016systran,michon2020integrating}.

These methods can effectively improve the terminology generation accuracy in textual terminology translation. However, they are not applicable to speech: constrained generation methods are time-consuming \cite{beurer2024guiding}, while data augmentation methods are difficult to implement due to modal differences.

\subsection{Speech Tasks with Terminology Intervention}
A common paradigm for speech tasks with terminology intervention involves three steps: first training for modal alignment, then retrieving terms via VectorDB, and finally adding the terms to prompts to intervene in SLM generation \cite{yang2024promptasr,feng2025enhancing}. The differences lie in the methods for modal alignment. For instance, Seal adopts contrastive learning \cite{sun2025seal}, while BR-ASR \cite{gong2025br} uses CLAP \cite{elizalde2023clap}. A few works don't follow this paradigm; Locate-and-Focus retrieves audio frames containing terms via semantic similarity, then performs replacement using Code-Switch-like techniques and additionally adds special tokens \cite{wu2025locate}.

These methods share the commonality of retrieving terms based on semantic similarity, which is simple and feasible. However, due to the rigor of terminology, we require every character to be completely accurate. Yet, semantic similarity methods tend to recall words semantically similar to the real terms, which is not the result we expect. Furthermore, all the aforementioned works are tested on general-domain datasets, which differ significantly from real-world terminology scenarios. Therefore, our proposed A2P method directly calculates the probability of term occurrence instead of relying on semantic similarity, and avoids scenario differences by training on terminology data.

\section{Method}

In this section, we commence with the formal problem formulation, followed by an introduction to the multi-modal feature extraction scheme. Subsequently, we detail the principles of the proposed A2P and its associated training loss function. We then elaborate on the curriculum learning strategy implemented during training. Finally, we delineate the operational workflow of the Retriever during inference. The overall architecture of the Retriever is shown in Fig. \ref{A2P}.

\subsection{Problem Formulation}
In terminology retriever tasks, we define the training data as $D_{\text{train}} = \{speech, src\_terms\}$, where $src\_terms$ refers to the terminology in the same language as the speech. During inference, the data is defined as $D_{\text{infer}} = \{speech, term\_bank\}$, where the $term\_bank$ contains both $src\_terms$ and their corresponding $tgt\_terms$. Our proposed A2P framework consists of four key stages: Multi-modal Feature Extraction, Cross-Modal Retriever Training, Curriculum learning, and Terminology-Augmented Inference. Each stage will be introduced sequentially in the following sections.
\input{figs/A2P}

\subsection{Multi-modal Feature Extraction}
There exists a modality gap between raw speech signals and textual data. To address this gap, we seek to align speech and text features within a shared latent space. The multimodal model Qwen2-Audio-Instruction exhibits the capacity for joint comprehension of both speech and textual modalities, indicating that its feature space inherently encodes representations of both. Accordingly, we utilize the audio encoder from Qwen2-Audio-Instruction to extract speech features, while textual inputs are directly processed via standard embedding layers. Our framework processes each modality through dedicated pathways:

\begin{align}
    \mathbf{h}_s &= f_{\theta}^{\text{audio}}(speech) \in \mathbb{R}^{d} \\
    \mathbf{h}_t &= g_{\phi}^{\text{text}}(src\_term) \in \mathbb{R}^{d}
\end{align}
where $f_{\theta}^{\text{audio}}$ and $g_{\phi}^{\text{text}}$ represents the parameter of Qwen2-Audio-Instruction's audio-encoder and embedding module. $d$-dimensional is the embedding dimension of Qwen2-Audio-Instruction. $f_{\theta}^{\text{audio}}$ and $g_{\phi}^{\text{text}}$ are kept frozen during training.

\subsection{Cross-Modal Retriever Training}
Inspired by advancements in the field of speech tasks with terminology intervention, we propose a speech retriever architecture. Using a cross-attention mechanism, our model computes the attention weights between speech signals and different candidate terms. These attention weights directly indicate the probability of each term's presence in the speech. The specific formula is as follows:
\begin{align}
    \mathbf{S}_{\text{attn}} &= \text{MHA}(\mathbf{h}_s, \mathbf{h}_t)
\end{align}
where $\text{MHA}$ is the multi-head cross attention, $\mathbf{h}_s$ and $\mathbf{h}_t$ are the feature of speech and terms.

While we have obtained attention weights between terms and speech signals, these weights remain inherently token-level due to the initial term processing through a tokenizer. However, our optimization objective requires computing the probability of term occurrence in the speech, which necessitates aggregating term-level information beyond token-level representations.

\input{figs/pooled}
To address this, we propose an effective pooling method: sum the token-level weights to obtain the overall attention weight of the speech towards terms. To ensure the stability of subsequent training, we normalize the attention weight based on the lengths of the speech and term, as illustrated in Figure \ref{pooled}. The specific formula is as follows:
\begin{align}
    \mathbf{S}_{\text{masked}} &= \mathbf{S}_{\text{attn}} \odot \mathbf{M}_{\text{speech}} \odot \mathbf{M}_{\text{text}} \label{eq:masked} \\
    \mathbf{S}_{\text{sum}} &= \sum_{i=1}^{L} \mathbf{S}_{\text{masked}}[i] \label{eq:token} \\
    \mathbf{M}_{\text{sum}} &= \sum_{i=1}^{L} \mathbf{M}_{\text{text}} \\
    \mathbf{S}_{\text{pooled}} &= \frac{\mathbf{S}_{\text{sum}}}{\mathbf{M}_{\text{sum}} + \epsilon} \label{eq:pooled}
\end{align}
where $\mathbf{M}_{\text{speech}}$ and $\mathbf{M}_{\text{text}}$ respectively represent the mask matrices of speech features and text features, and $\epsilon$ is a minimal number used to avoid the situation where the denominator is zero. 

First, we extract the relevant signals using mask matrices to eliminate irrelevant positions in Eq. \ref{eq:masked}. Subsequently, summation along the term-length dimension utilizes the actual term length to prevent gradient instability caused by high-dimensional interactions, where division by the term length serves as dimensional reduction regularization. This design explicitly converts fine-grained token correlations into robust term-level representations while maintaining numerical stability throughout the training process.
In the end, we implement a residual connection \cite{he2016deep} to preserve original semantic information.
\begin{align}
    \mathbf{S}_{\text{final}} &= \mathbf{h}_t + \mathbf{S}_{\text{pooled}} \\
    \hat{y} &= \sigma(Linear(\mathbf{S}_{\text{final}}))
\end{align}

To address data scarcity in speech-to-text tasks with terminology intervention, we calculate the loss of positive and negative samples simultaneously. This helps the model learn to recognize terms present in the speech while distinguishing absent terms, thereby improving retrieval robustness. The formalized loss function combines dual objectives:
\begin{align}
\mathcal{L} = \underbrace{\mathbb{E}{(s,t^+)}[-\log\hat{y}^+]}_{\text{Positive term maximization}} + \underbrace{\mathbb{E}{(s,t^-)}[-\log(1-\hat{y}^-)]}_{\text{Negative term minimization}}
\end{align}
Where $\hat{y}^+$/$t^+$ denote positive samples (term present in speech) and $\hat{y}^-$/$t^-$ represent negative samples (term absent). 

\subsection{Curriculum Learning}
Due to the highly heterogeneous length distribution of real-world terms, a retriever without well-initialized parameters struggles to effectively learn both short and long terms simultaneously. To tackle this training challenge, we adopt a curriculum learning strategy \cite{bengio2009curriculum, wang2021survey}, by organizing the training process into three successive stages:
\begin{itemize}
\item Word-level: In this stage, terms within the training data consist of randomly selected individual words from the text. The fine-grained nature and uniform length of these terms significantly reduce the learning difficulty of A2P, facilitating initial parameter estimation.
\item Phrase-Level: Building upon the acquired ability to detect fine-grained terms, this stage utilizes randomly selected sequences of 1 to 4 consecutive words within the text as terms. The objective is to progressively enhance the A2P sensitivity in terms of varying lengths.
\item Real-Term Level: Following the preceding stages, the retriever possesses well-initialized parameters. Consequently, this final stage employs authentic terms for training, ensuring robust recall capability under real-world conditions.
\end{itemize}

\subsection{Terminology-Augmented Inference}
During deployment, the trained retriever performs efficient cross-modal matching between input speech and a predefined terminology bank $term\_bank = {t}_{i=1}^m$ containing $m$ terms. The retrieval process is formalized as:
\begin{align}
p(t_i|speech) &= \sigma(\psi_\theta(speech, t_i)), \quad \forall i \in [0, m] \label{eq:retri} \\
\mathcal{K} &= \underset{1 \leq i \leq m}{\text{argtopk}} \left(p(t_i|speech)\right) \\
T_{\text{retrieved}} &= {term\_bank[k], \quad \forall k \in \mathcal{K}}
\end{align}
where $\psi_\theta$ represents the retriever parameters, Eq. \ref{eq:retri} is used to compute the presence probability of terms from the terminology base within the speech, utilizing the Retriever. These terms are then ranked by descending probability, and the indices of the Top-K terms are retrieved. These indices query the terminology base to obtain the candidate term set.

Upon retrieving candidate terms $T_{\text{retrieved}}$, we spell them into the prompt and inject this knowledge by leveraging the ICL capabilities of LLM.

\begin{tcolorbox}[boxrule=1pt,boxsep=1pt,left=2pt,right=2pt,top=2pt,bottom=2pt]
\textbf{Prompt Example:}
This is an English audio recording. Please transcribe this audio into English text. Specialized terminology may appear in the audio. Please accurately recognize these terms. Potential technical terms include: SARS, RSA, SARS-CoV, Boris, SARS-CoV-2, respiratory, respirator, Tapia, war, and respiratory disease.
\end{tcolorbox}

\input{tables/data}
\section{Experiments}
\subsection{Dataset}
To date, there are no open-source datasets for speech-to-text tasks with terminology intervention. Even datasets for textual terminology translation are extremely scarce. Therefore, we devise an alternative strategy: 

We repurpose entities from Named Entity Recognition (NER) datasets as terms, and then utilize MegaTTS to generate corresponding speech for their textual definitions. As shown in Table \ref{example1datalist}, we collect data from both Chinese and English, and also evaluate the performance of these two language directions in subsequent tasks. Such as Wikiann \cite{pan2017cross}, MSRA-NER \cite{levow2006third}, Few-nerd \cite{ding2021few} and CMeEE \cite{du2022mrc}.

Given the limited data volume achievable via the first strategy, we further augment our data by leveraging the LibriSpeech \cite{panayotov2015librispeech} and Aishell-2 \cite{du2018aishell} datasets, designating certain words or phrases within their utterances as terms. This approach enables the cost-effective acquisition of large-scale data.

To evaluate model performance, we utilize the Chinese-English terminology translation test set from the machine translation dataset ComMT \cite{luo2025beyond}. This high-confidence dataset comprises carefully curated terms sourced from authoritative benchmarks, including WMT and DAS, with domain coverage concentrated in news and medicine. Consistent with the prior experimental configuration, we also synthesize bidirectional speech corpora using MegaTTS.

\subsection{Experimental setups}
\subsubsection{Pre-processing} For speech data, LibriSpeech has a sampling rate of 16 kHz, while other TTS-generated speech data use a sampling rate of 24 kHz. During speech loading, all data is uniformly resampled to 16 kHz. For terminology data, the word-level terms in LibriSpeech are sourced from the all\underline{ }rare subset in Rare5k \cite{le2021contextualized}, while phrase-level terms are constructed by randomly selecting text spans from utterances. For Aishell-2, words or phrases are randomly selected to facilitate both word-level and phrase-level training. Terms in TTS-generated data correspond to entity names from the original NER datasets. All data undergoes feature extraction strictly following the default configuration of Qwen2-Audio-Instruction, with the original vocabulary retained without modification.

\subsubsection{Model Setting}
We employ the Qwen2-Audio-Instruction model as our SLM. The instruction-tuned variant is selected because terms are dynamically incorporated into the prompt, necessitating robust instruction-following capabilities, particularly in handling dynamically constructed prompts. For the audio encoder, we utilize the pretrained weights from Qwen2-Audio-Instruction. To ensure compatibility with the encoder, the hidden dimension of the cross-attention retriever is accordingly set to 4096. The Retriever employs only a single-layer cross-attention mechanism, with the multi-head attention configured with 32 heads and a dropout rate of 0.1. 
\subsubsection{Retriever Training and Inference}
During training, the batch size is set to 32, with a maximum term bank capacity of 100 terms per batch. We train with a maximum of 50 epochs and a learning rate of 1e-4. The initial learning rate is 1e-7, accompanied by 500 warmup steps. The CosineAnnealingLR scheduler is adopted for learning rate decay. Optimization used AdamW with hyperparameters $\beta_1$=0.9,  $\beta_2$= 0.98, and weight decay = 0.01.

During inference, the Retriever operates on a terminology bank comprising all 583 terms from the test set. All models are trained on 8 Nvidia Tesla A100-80G GPUs. We implement our models based on the Accelerator.

\input{tables/recalls}

\subsubsection{Instruction Fine-Tuning}
When intervening in ASR and ST tasks with terminology, robust instruction-following capabilities become essential for the SLM due to the added terminology. To ensure effective terminology utilization, we perform instruction fine-tuning on Qwen2-Audio-Instruction.

The Niutrans\footnote{challenge.xfyun.cn/topic/info?type=machine-translation-2024} dataset from Table \ref{example1datalist}, containing parallel English-Chinese corpora, is used for fine-tuning. To enhance the SLM's instruction adherence, ASR and ST tasks are randomly interleaved during training. Furthermore, terminology presentation is randomly assigned to four conditions: no terminology, entirely correct terminology, entirely incorrect terminology, and partially correct terminology. 

We use the LlamaFactory toolkit for training, with 3 epochs and a learning rate of 2e-6 \cite{zheng2024llamafactory}. We employ full fine-tuning within the DeepSpeed framework \cite{rajbhandari2020zero}, setting the stage parameter to 3.

\subsubsection{Baseline Systems}
To validate the effectiveness of our approach, we conduct comparative experiments against a conventional VectorDB solution. The Chroma VectorDB is employed, utilizing cosine similarity for distance computation, with the SONAR modality alignment model selected as the SOTA open-source baseline.
We adopt the HNSW algorithm for indexing, while other parameters are configured using Chroma's default settings. 

To further investigate the impact of the audio encoder on Retriever performance, we evaluate three alternative encoders: Qwen-Audio-Chat and SONAR. Notably, as SONAR inherently outputs dense embeddings, pooling operations are omitted for the specific encoder.

\section{Results and Analysis}
In this section, we will first present the recall rate of Retriever under different Settings and analyze the performance differences between A2P and VectorDB. Next, we will analyze the impact of curriculum learning on the Retriever. After that, we analyze what effects different Audio-Encoders have on the Retriever and analyze the reasons. Then, we will analyze the changes in recall performance and recall speed under different sizes of terminology databases. Last but not least, the changes in LLM performance before and after using recall terms to intervene in speech recognition and speech translation will be demonstrated.

\subsection{Main Results of the A2P Method}
Table \ref{recalls} presents the recall rates of various approaches under multiple conditions. It is observable that the A2P approach achieves high terminology recall accuracy with specific audio encoders, surpassing the VectorDB method by a substantial margin. A2P exhibits an absolute performance advantage, particularly when retrieving fewer terms.

We conduct an analysis to understand this phenomenon, revealing an intriguing insight: the VectorDB method fundamentally doesn't perform retrieval; instead, it identifies semantically similar vectors within the embedding space.

\subsubsection{Core Limitation of VectorDB Approach}
The VectorDB finds the term vectors closest to the speech vector using cosine similarity. The similarity between terms and speech in the semantic space \textbf{does not equal} the probability of the term occurrence in the speech. This occurs because compressing the audio into a single vector makes the semantic information highly concentrated. As a result, during retrieval, the VectorDB often recalls terms that are semantically relevant but not present in the original audio.

\subsubsection{Advantages of of A2P} The A2P method is explicitly optimized to detect terms that \textbf{are present} within the speech signal. This optimization objective results in retrieved terms having a significantly higher probability of actual occurrence in the speech.

\input{tables/speed}

\subsection{The Impact of Audio-Encoder}
Further Analysis from Table \ref{recalls}: A pronounced performance discrepancy is observed across different audio encoders. When employing SONAR as the audio encoder, the A2P demonstrates suboptimal performance on the English test set and fails to yield measurable results on the Chinese test set. We attribute this limitation to SONAR extraction of ultra-high-dimensional features, which exceeds the modeling capacity of our lightweight A2P architecture (80M parameters). Scaling the Retriever to parameter magnitudes comparable to SeamlessM4T (e.g., billion-scale) might enable effective utilization of SONAR feature representations.

Qwen-Audio-Chat exhibits disparate performance: it approaches VectorDB performance on Chinese but underperforms on English.  We hypothesize this stems primarily from imbalances in its training data, likely reflecting greater Chinese data abundance during training. 
These findings address the question posed in the Introduction. Firstly, modality alignment proves critical for cross-modal retrieval, directly impacting performance. Secondly, some SLMs inherently possess this alignment capability without requiring dedicated training. The existing Qwen2-Audio-Instruction model already demonstrates strong inherent alignment.

\subsection{Different Sizes of Terminology Databases}
\subsubsection{Speed} Table \ref{speed} presents the average retrieval time per query for both methods. The VectorDB solution consistently outperforms the A2P approach across all tested terminology database scales. Further analysis of A2P's latency reveals that approximately 60\% of the processing time is consumed by the Top-k selection algorithm. Top-k operation, with its $O(N \log K)$ time complexity, represents the primary computational bottleneck in the A2P pipeline.

Furthermore, it can be clearly observed from Table \ref{speed} that the latency of A2P increases with the expansion of the vocabulary size, which is attributed to the increased computational load. By contrast, the retrieval speed of VectorDB remains consistently stable. Moreover, we argue that the retrieval speed can be further enhanced by adopting more efficient indexing methods (e.g., GPU-optimized indexes), a performance that A2P cannot match.

\input{figs/sizebank}
\input{tables/course-learning}
\input{tables/s2t}
\subsubsection{Recall} Within the same language, A2P consistently outperforms the VectorDB approach across all metrics, as shown in Fig. \ref{fig5}. Furthermore, while A2P maintains robust performance on Chinese, it exhibits a more pronounced decline on English. We attribute this divergence to fundamental differences in language typology.

The typological distinction between Chinese (an isolating language) and English (a fusional language) leads to drastically different semantic representations \cite{wiltschko2008syntax}. This manifests in terminology retrieval: terms phonetically similar to English roots or affixes are highly susceptible to false recalls. As the database grows, such false recalls become more prevalent, consequently causing significant performance degradation in English.

\subsection{Ablation Study}
Table \ref{course-learning} presents an ablation study on the impact of curriculum learning stages on the A2P final performance. Key observations are as follows:
\begin{itemize}
    \item Directly training the A2P on real-term datasets during the initial phase yields no measurable performance.
    \item When trained solely at the word-level, the Retriever demonstrates limited recall capability. However, recall precision remains suboptimal, with retrieved terms exhibiting low diversity and a pronounced bias toward short words.
    \item Augmenting training with phrase-level data significantly improves performance, and the retriever shows the ability to recall the right terms.
    \item Subsequent fine-tuning on real-term datasets after phrase-level training produces a substantial enhancement in overall recall performance.
\end{itemize}

These experiments indicate that authentic data constitutes the critical success factor for the Retriever, while word-level and phrase-level training primarily mitigate data scarcity limitations. Our generated audio corpus ($\sim$100k samples) likely falls below the scaling law threshold required for optimal Retriever performance. Consequently, a Curriculum Learning training pipeline may become unnecessary when sufficient authentic data is available.

Furthermore, ablating the token-level pooling drastically reduced the overall recall of A2P to below 1\%. We attribute this to inherent redundancy in token-level attention weights. The simple linear structure struggles to capture the holistic semantics of terms from this complex information. The result strongly validates the design rationale behind the A2P architecture.

\subsection{The Influence of A2P on ASR and ST}
Table \ref{s2t} presents the performance changes for ASR and ST tasks after terminology intervention using Qwen2-Audio-Instruction. The A/B metric represents text quality versus terminology generation accuracy. For ASR, Character Error Rate (CER) is used for Chinese evaluation, and WER for English. For ST, BLEU scores are used for both languages.

Terminology generation accuracy showed significant improvement across both tasks. For ASR, accuracy increased by approximately 6\% on average in English and 5\% in Chinese. The improvement is more pronounced in ST tasks, with English-to-Chinese translation gaining about 12\% and Chinese-to-English improving by 13\%.

While the intervention approach achieves strong performance in ASR, limitations remain in ST. Notably, intervention improves both BLEU scores and terminology accuracy in ST, yet the terminology accuracy remains unsatisfactory. The complexity of ST for SLM contributes to the terminology accuracy being significantly lower than the recall rates in ST. This indicates that many correctly retrieved terms are not effectively utilized by the SLM. However, these results also demonstrate the challenging nature of our proposed dataset for current SLMs.

Furthermore, SLM performance doesn't strictly improve with increasing terminology volume. For example, increasing terms from 30 to 40 degraded performance in English ASR but improved it in Chinese ASR. We attribute this phenomenon primarily to limitations in the SLM's ICL capabilities. Insufficient ICL capabilities of SLM likely hinder maintaining ASR/ST performance when processing numerous terms.

These findings reveal new challenges for speech-to-text tasks with terminology intervention. 

(1) How can terminology accuracy in ST tasks be effectively increased? 

(2) How can SLMs preserve baseline ASR/ST performance when handling multiple terms?

\section{Conclusion}
This paper first addresses the data scarcity issue in terminology-augmented ASR and ST tasks by generating a labeled dataset using open-source MT, NER data, and MegaTTS. The public release of this dataset is expected to significantly advance research in this domain. Subsequently, to mitigate terminology retrieval accuracy limitations, we propose the A2P methodology, which effectively circumvents two critical constraints inherent in VectorDB approaches while achieving superior recall rates in experimental evaluations. Notwithstanding these advancements, A2P currently exhibits high latency compared to VectorDB in retrieval throughput. Crucially, we demonstrate that systematic incorporation of retrieved terminology into SLM architectures significantly enhances terminology generation accuracy across both ASR and ST tasks. Additionally, our analysis reveals two critical issues in existing SLMs, offering valuable insights for future research in this field.

\bibliography{aaai2026}

\end{document}

%% file: figs/overview.tex
\begin{figure}[t]
\centering
\centering
\includegraphics[width=0.9\columnwidth]{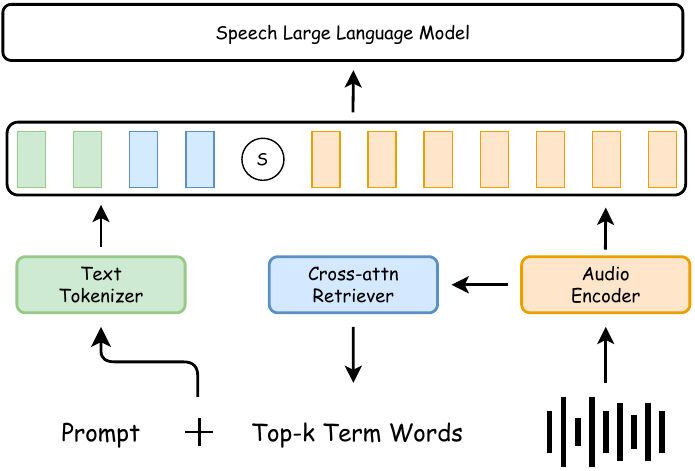}
\caption{The overall architecture of Attention2Probability. Audio features are extracted and then fed into a cross-attention retriever, which retrieves the Top-k terms with the highest probability of occurrence within the audio. These retrieved terms are concatenated with the prompt. Finally, the prompt and the audio features are jointly input into the speech large language model.}
\label{overview}
\end{figure}

%% file: figs/A2P.tex
\begin{figure}[t]
\centering
\centering
\includegraphics[width=0.9\columnwidth]{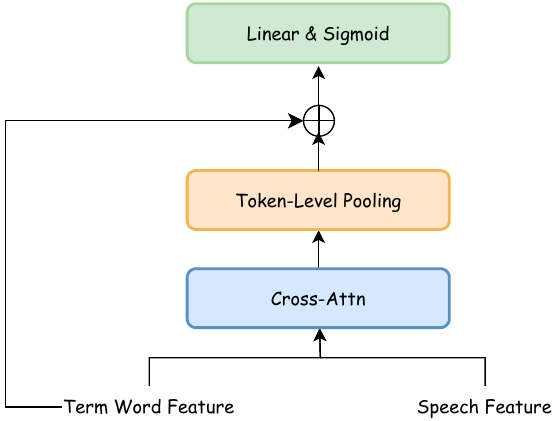}
\caption{The overall architecture of the Retriever in the A2P method.}
\label{A2P}
\end{figure}

%% file: figs/pooled.tex
\begin{figure}[t]
\centering
\centering
\includegraphics[width=0.9\columnwidth]{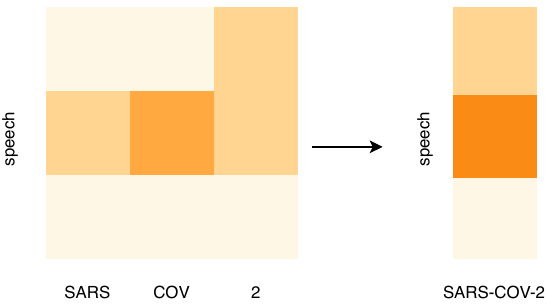}
\caption{Schematic of token-level pooling. For example, a multi-word term like "SARS-COV-2" is tokenized by the tokenizer into subword units. Token-level pooling aggregates these subword representations, thus preserving the term's semantic integrity and allowing the retriever to focus on the holistic information of the term.}
\label{pooled}
\end{figure}

%% file: tables/data.tex
\begin{table}[t]
\centering
\small
\begin{tabular}{L{1.7cm}C{1.2cm}C{1.2cm}C{1.5cm}}\toprule
Name      &  Hours  & Sents & Language \\\midrule
ComMT    &  7.5 & 3.3K         & EN, ZH           \\
Niutrans   &  36 & 18.5K         & EN, ZH           \\
Librispeech  &  960 & 286.8K         & EN            \\
Few-nerd & 177 & 58.5K & EN \\
Wikiann  & 13 & 11.5K & EN \\
Kaggle-NER  &  106 & 40.9K         & EN            \\
Aishell-2 & 1000 &560.5K & ZH \\
MSRA-NER & 80 &27.5K & ZH \\
NLPC2018 & 12 &13.1K & ZH\\
CMeEE & 60  & 20K& ZH\\\bottomrule
\end{tabular}

\caption{Statistics of all datasets.}
\label{example1datalist}
\end{table}

%% file: tables/recalls.tex
\begin{table*}[t]
\centering
\small

\begin{tabular}{C{1.3cm}L{1.8cm}C{3.3cm}C{1.4cm}C{1.4cm}C{1.4cm}C{1.4cm}C{1.4cm}}\toprule
Language &  Retriever type  & Audio-encoder & Top-10 & Top-20 & Top-30 & Top-40 & Top-50 \\\midrule
\multirow{4}{*}{EN} & VectorDB & SONAR & 62.89 & 73.74 & 77.98 & 81.11 & 83.49\\
  & A2P & SONAR & 15.15 & 26.57 & 30.38 & 31.58 & 33.93\\
   & A2P & Qwen-Audio-Chat & 7.04 & 13.15 & 17.64 & 21.71 & 25.27\\
    & A2P (Ours) & Qwen2-Audio-Instruction & \textbf{75.55} & \textbf{81.57} & \textbf{83.82} & \textbf{85.72} & \textbf{86.83}\\\bottomrule\addlinespace[3pt]
    \multirow{4}{*}{ZH} & VectorDB & SONAR & 58.46 & 67.22 & 72.51 & 78.19 & 81.51\\
  & A2P & SONAR & - & - & - & - & -\\
   & A2P & Qwen-Audio-Chat & 60.32 & 69.74 & 73.69 & 76.15 & 78.01 \\
    & A2P (Ours) & Qwen2-Audio-Instruction & \textbf{83.31} & \textbf{89.44} & \textbf{91.03} & \textbf{92.29} & \textbf{92.57}\\\bottomrule
\end{tabular}
\caption{The recall of Retriever under different Settings. - means less than 1\%}
\label{recalls}
\end{table*}

%% file: tables/speed.tex
\begin{table}[t]
\centering
\small

\begin{tabular}{L{1.3cm}C{1.2cm}C{1.2cm}C{1.2cm}C{1.2cm}}\toprule
      &  583 & 1k & 5k & 10k  \\\midrule
  A2P &  8.71ms & 13.44ms & 62.86ms& 126.31ms\\
VectorDB & 0.74ms & 0.85ms & 1.17ms & 0.91ms \\\bottomrule
\end{tabular}
\caption{Recall speed under different terminology databases.}
\label{speed}
\end{table}

%% file: figs/sizebank.tex
\definecolor{color1}{RGB}{255,000,000}
\definecolor{color2}{RGB}{046,117,182}
\definecolor{color3}{RGB}{112,173,071}
\definecolor{color4}{RGB}{237,125,049}
\definecolor{color5}{RGB}{158,072,014}
\definecolor{color6}{RGB}{165,165,165}
\definecolor{color7}{RGB}{255,192,000}
\definecolor{color8}{RGB}{112,048,160}
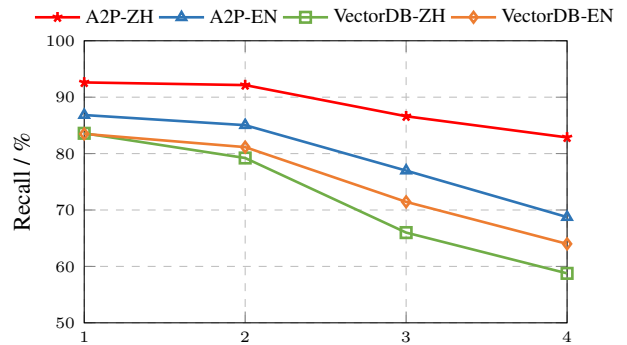
\begin{figure}[t]
\centering
  \begin{tikzpicture}
  \tikzstyle{textonly} = [font=\scriptsize,align=left]
  \node[textonly] at (5,0) {};
      \begin{axis}[
        fill opacity=1,
        fill=orange,
        ymajorgrids,
        xmajorgrids,
        grid style=dashed,
        width=0.45\textwidth,
        height=0.3\textwidth,
        legend columns=-1,
        legend entries={
         A2P-ZH, A2P-EN, VectorDB-ZH, VectorDB-EN
        },
        legend style={fill opacity=0.5,text opacity =1,
          draw=none,
          line width=1pt,
        },
        legend style={
        at={(0.5,1.15)}, anchor=north,
        nodes={scale=0.75, transform shape}
        },
        xmin=1, xmax=4,
        ymin=50, ymax=100,
        xtick={1, 2, 3, 4},
        ytick={50,60,70,80,90,100},
        xticklabel style={font=\tiny},
        yticklabel style={font=\tiny},
        ylabel=\footnotesize{Recall / \%},
        ylabel style={yshift=-0.1em},
        xlabel style={yshift=0.1em},
        scaled ticks=false,
        ]

    \addplot[color1,mark=star, line width=1pt] 
    file {data/A2P_zh.txt}; 
    \addplot[color2,mark=triangle,line width=1pt]
    file {data/A2P_en.txt};
    \addplot[color3,mark=square, line width=1pt] 
    file {data/vectorDB_zh.txt};
    \addplot[color4, mark=diamond, line width=1pt] 
    file {data/vectorDB_en.txt};  
    \end{axis}
    \end{tikzpicture}
\caption{The trend of the recall rate when k=50 as the terminology database size increases is shown below. Here, 1 denotes 583 terms, while 2, 3, and 4 denote 1k, 5k, and 10k terms, respectively.}
\label{fig5}
\end{figure}

%% file: tables/course-learning.tex
\begin{table*}[t]
\centering
\small

\begin{tabular}{L{3.6cm}C{1.3cm}C{1.3cm}C{1.3cm}C{1.3cm}C{1.3cm}}\toprule
 & Top-10 & Top-20 & Top-30 & Top-40 & Top-50 \\\midrule
A2P & 75.55 & 81.57 & 83.82 & 85.72 & 86.83\\
\hspace{1em}- token-level pooling & - & - & - & - & -\\
\hspace{1em}- word / pharse-level & - & - & - & - & -\\
\hspace{1em}- real-term & 42.50 & 55.31 & 61.69 & 65.73 & 69.22 \\
\hspace{2em}- pharse-level & 27.05 & 39.79 & 46.62 & 51.16 & 54.73 \\
\hspace{2em}- word-level & - & - & - & - & - \\
\bottomrule
\end{tabular}
\caption{Results of the ablation experiment. - means less than 1\%}
\label{course-learning}
\end{table*}

%% file: tables/s2t.tex
\begin{table*}[t]
\centering
\small

\begin{tabular}{C{1.1cm}C{1.1cm}C{1.5cm}C{1.5cm}C{1.5cm}C{1.5cm}C{1.5cm}C{1.5cm}}\toprule
 & & Top-0 & Top-10 & Top-20 & Top-30 & Top-40 & Top-50 \\\midrule
\multirow{2}{*}{ASR} & EN & 12.29/79.66 & 11.44/85.90 & 11.52/85.81 & 11.27/86.09 & 11.48/85.11 & 11.50/85.65 \\
 & ZH & 13.55/83.31 & 11.91/91.25 & 10.91/91.62 &  10.73/88.81 & 10.12/90.69 & 10.32/90.48\\\bottomrule\addlinespace[3pt]
\multirow{2}{*}{ST} & EN-ZH & 28.73/53.47 & 32.32/67.93 & 31.95/65.81 & 32.21/66.58 & 31.51/65.91 & 32.41/66.12 \\
 & ZH-EN & 16.57/46.42 & 18.61/65.58 & 18.69/64.69 & 18.40/65.04 & 18.60/64.07 & 18.95/63.19 \\\bottomrule
\end{tabular}
\caption{ASR and ST Task Performance with Terminology Intervention. }
\label{s2t}
\end{table*}